\documentclass[10pt,twocolumn,letterpaper]{article}

\usepackage[pagenumbers]{cvpr} 

\definecolor{cvprblue}{rgb}{0.21,0.49,0.74}
\usepackage[pagebackref,breaklinks,colorlinks,allcolors=cvprblue]{hyperref}
\usepackage{lineno}
\def\paperID{1403} 
\def\confName{CVPR}
\def\confYear{2025}

\usepackage{enumitem}

\usepackage{booktabs}
\usepackage{graphicx}
\usepackage{float,wrapfig}
\usepackage{amsmath}
\usepackage{amsfonts}
\usepackage{amssymb}
\usepackage{arydshln}
\usepackage{pifont}

\usepackage[pagebackref,breaklinks,colorlinks]{hyperref}

\usepackage{array}
\usepackage{adjustbox}
\usepackage{multirow}
\usepackage{colortbl}
\usepackage{caption}
\usepackage{amsmath}
\usepackage{multicol}
\usepackage{mathtools}

\usepackage{soul}
\usepackage{makecell}

\usepackage{color,xcolor}

\usepackage{bm}

\usepackage{url}
\usepackage{algorithm, algorithmic}
\usepackage[super]{nth}

\makeatletter
\DeclareRobustCommand\onedot{\futurelet\@let@token\@onedot}
\def\@onedot{\ifx\@let@token.\else.\null\fi\xspace}

\makeatother

\usepackage[capitalize]{cleveref}
\crefname{section}{Sec.}{Secs.}
\Crefname{section}{Section}{Sections}
\Crefname{table}{Table}{Tables}
\crefname{table}{Tab.}{Tabs.}

\newcommand{\method}{MultiMorph}

\definecolor{rowblue}{RGB}{220,230,240}
\definecolor{myorchid}{RGB}{150,10,30}
\definecolor{myblue}{RGB}{10,30,250}
\definecolor{mygreen}{RGB}{10,120,10}

\def\eqref#1{equation~\ref{#1}}

\def\1{\bm{1}}

\def\vt{{\bm{t}}}
\def\vu{{\bm{u}}}

\def\vx{{\bm{x}}}

\DeclareMathAlphabet{\mathsfit}{\encodingdefault}{\sfdefault}{m}{sl}
\SetMathAlphabet{\mathsfit}{bold}{\encodingdefault}{\sfdefault}{bx}{n}

\newcommand{\R}{\mathbb{R}}

\DeclareMathOperator*{\argmin}{arg\,min}

\newcommand{\subpara}[1]{\vspace{0.4em} \noindent \textbf{#1}}

\begin{document}

\title{\method{}: On-demand Atlas Construction}

\author{S. Mazdak Abulnaga$^{1,2}$ \qquad Andrew Hoopes$^{1,2}$ \qquad Neel Dey$^{1,2}$ \qquad Malte Hoffmann$^{2}$ \\
Marianne Rakic$^{1,2}$ \qquad Bruce Fischl$^{2}$ \qquad John Guttag$^{1}$ \qquad Adrian Dalca$^{1,2}$ \\
{\small $^{1}$MIT Computer Science and Artificial Intelligence Laboratory} \qquad {\small $^{2}$Massachusetts General Hospital, Harvard Medical School} \\
    {\small \texttt{abulnaga@csail.mit.edu}}
}

\maketitle

\begin{abstract}
    \looseness=-1
    We present \method{}, a fast and efficient method for constructing anatomical atlases on the fly. Atlases capture the canonical structure of a collection of images and are essential for quantifying anatomical variability across populations.
    However, current atlas construction methods often require days to weeks of computation, thereby discouraging rapid experimentation.
    As a result, many scientific studies rely on suboptimal, precomputed atlases from mismatched populations, negatively impacting downstream analyses. 
    \method{} addresses these challenges with a feedforward model that rapidly produces high-quality, population-specific atlases in a single forward pass for any 3D brain dataset, without any fine-tuning or optimization. 
    \method{} is based on a linear group-interaction layer that aggregates and shares features within the group of input images.
    Further, by leveraging auxiliary synthetic data, \method{} generalizes to new imaging modalities and population groups at test-time. 
    Experimentally, \method{} outperforms state-of-the-art optimization-based and learning-based atlas construction methods in both small and large population settings, with a 100-fold reduction in time.
    This makes \method{} an accessible framework for biomedical researchers without machine learning expertise, enabling rapid, high-quality atlas generation for diverse studies.

\end{abstract}
\section{Introduction}
\label{s:intro}
We present \method{}, a rapid and flexible method for constructing anatomical atlases. An atlas, or deformable template, is a reference image that represents the typical structure within a collection of related images. In  biomedical imaging studies, atlases facilitate studying anatomical variability within and across population groups by serving as as a common coordinate system for key image analysis tasks such as segmentation~\cite{van1999automated,fischl2002whole,grau2004improved,ashburner2005unified}, shape analysis~\cite{felzenszwalb2005representation,kokkinos2012intrinsic,monti2017geometric,alp2017densereg}, and longitudinal modeling~\cite{reuter2011avoiding,reuter2012within,hoffmann2020long}.


\looseness=-1
Traditional unbiased atlas construction for a population involves solving a computationally intensive iterative optimization problem that often requires several days or weeks of computation. The optimization alternates between aligning (registering) all images to the estimated atlas and updating the atlas in both shape and appearance by averaging the images mapped to the intermediate atlas space~\cite{joshi2004unbiased,avants2010optimal}. 
Recent learning-based methods employ a target dataset to explicitly learn an atlas jointly with a registration model~\cite{dalca2019learning,dey2021generative}, yet still require days of training. This necessitates computational infrastructure and machine learning expertise that is unavailable to many biomedical researchers.

\looseness=-1
Regardless of strategy, an atlas produced from one population of images may not be appropriate for populations that differ from the group used to build the atlas. Re-estimating the atlas is often required for each new experiment. These computational challenges are further compounded by the need to construct atlases for specific image types as many biomedical studies acquire several imaging modalities to highlight different biomedical properties of interest. The repeated, prohibitive computational cost of producing a new atlas leads most scientists to use existing atlases that might not be appropriate for their population group or modality, thereby negatively impacting the analyses in these studies~\cite{lancaster2007bias}.



To meet these challenges, we introduce \method, a machine learning model that constructs atlases in a single forward pass, requiring only seconds to minutes of computation on a CPU, and no machine learning expertise to use. \method{} efficiently generates population- and subgroup-specific atlases, enabling accurate and fine-grained anatomical analyses. We employ a convolutional architecture that processes an arbitrary number of images and computes a set of regularized deformation fields that align the group of images to an atlas space central to that group. The proposed method uses a nonparametric convolutional operation that interacts the intermediate representations of the input images with each other, summarizing and aggregating shared features. Further, by training on diverse imaging modalities alongside supplementary synthetic neuroimaging volumes~\cite{gopinath2024synthetic}, \method{} generalizes to arbitrary imaging modalities at test time. We also introduce a \textit{centrality} layer that ensures that the estimated atlases are unbiased~\cite{joshi2004unbiased}. As a result, \method{} rapidly produces high quality atlases for new populations and imaging modalities unseen during training. 
It further yields more accurate segmentation transfer across population groups than both the most widely used optimization-based approach~\cite{avants2010optimal} and recent machine learning approaches~\cite{dalca2019learning,ding2022aladdin}. To summarize our contributions:
\begin{itemize}
    \item We frame atlas construction as a learning-based group registration problem to a central space.
    \item We present a novel neural network architecture that enables communication between the intermediate representations of a group of images, and show how this can be used to construct accurate group-specific atlases.
    \item We develop a \textit{centrality} layer that encourages predicted deformations and atlases to be central and unbiased.
    \item Experimentally, \method{} produces atlases that are as good, and often better, than those produced by other methods--and it does it up to 100 times faster. 
    \item We demonstrate the generalizability of the proposed method by constructing atlases for unseen imaging modalities and population groups. These atlases conditioned on age and disease state capture population trends within the data, enabling cross-group analyses.
    
\end{itemize}

\noindent Our model weights and code are available at \url{https://github.com/mabulnaga/multimorph}.

\section{Related work}

\paragraph{Deformable Registration.}
Deformable registration estimates a dense spatial mapping between image pairs. Traditional methods~\cite{rueckert1999nonrigid,ashburner2007fast,rohr2001landmark,modat2010fast,avants2008symmetric,siebert2024convexadam,jena2024fireants} solve an optimization problem balancing image-similarity 
and regularization terms to ensure smooth, invertible deformations. 

\looseness=-1
Learning-based methods improve test-time efficiency by training models to directly predict transformations between image pairs, generally enabling faster predictions on new image pairs as compared to traditional methods. Supervised approaches~\cite{sokooti2017nonrigid,rohe2017svf,yang2017quicksilver,young2022superwarp} are trained to regress simulated deformations or the outputs of registration solvers, whereas unsupervised methods~\cite{de2017end,balakrishnan2019voxelmorph,dalca2019unsupervised,krebs2019learning,hoffmann2021learning,hoffmann2021synthmorph,wang2023robust,hoffmann2024anatomy,gopinath2024registration,mok2023deformable,su2023nonuniformly,grzech2022variational,meng2023non,zhao2023spineregnet,chang2023cascading,qiu2023aeau} optimize an unsupervised image-similarity loss and a regularization term in training. 

\subpara{Synthetic Data in Neuroimage Analysis.} Recent machine learning-based neuroimage analysis methods have benefited from synthetic training data that extend far beyond real-world variations~\cite{billot2020partial,hoffmann2021learning,gopinath2024synthetic,hoopes2022synthstrip,kelley2024boosting,billot2023robust,billot2023synthseg,hendrickson2023bibsnet,shang2022learning,hoffmann2021synthmorph,hoffmann2024anatomy,iglesias2023ready}.
%
This domain-randomization strategy trains neural networks on simulated intensity images, synthesized on the fly from a training set of anatomical segmentation maps. As part of the generative model, the images undergo corruption steps simulating common acquisition-related artifacts like distortion~\cite{andersson2003correct}, low-frequency intensity modulation~\cite{sled1998nonparametric}, global intensity exponentiation~\cite{huang2012efficient}, resolution reduction, partial voluming~\cite{van2003unifying}, among many others.
The large variety of data yields shape-biased networks agnostic to the imaging modality.
As a result, these models generalize to arbitrary medical images that have the same anatomy as the synthetic training data -- largely eliminating the need for retraining to maintain peak performance~\cite{hoffmann2023anatomy,dey2024learning}. 


\subpara{Atlas Construction.} Deformable atlas construction seeks to find an image that optimally represents a given population, for example, to facilitate atlas-based brain segmentation~\cite{van1999automated,fischl2002whole,grau2004improved,ashburner2005unified} or to initialize longitudinal morphometric analyses in an unbiased fashion~\cite{reuter2011avoiding,reuter2012within,hoffmann2020long}.

Iterative atlas construction alternates between registering each image of the population to a current estimate of the atlas and updating the atlas with the average of the moved images until convergence~\cite{joshi2004unbiased,ma2008bayesian,avants2010optimal,sawiak2013voxel,pauli2018high}. Another approach computes a spanning tree of pair-wise transforms between subjects to estimate an atlas~\cite{seghers2004construction,iglesias2018model}. Iterative methods on $3$D data incur prohibitively long runtimes due to the cost of optimization. Therefore, many studies have used publicly available atlases~\cite{fonov2009unbiased}, although these are often not representative of the population being studied. 

Recent learning-based atlas construction techniques jointly learn an atlas and a registration network that maps images from the training population to the atlas~ \cite{dalca2019learning,dey2021generative,ding2022aladdin,sinclair2022atlas,starck2024diff,chi2023dynamic,grossbrohmer2024sina,dannecker2024cina}. These approaches naturally extend to constructing \textit{conditional} atlases, for example conditioning on age~\cite{dalca2019learning,dey2021generative,starck2024diff}, or incorporating tasks like segmentation~\cite{sinclair2022atlas}. However, obtaining an atlas for a new population requires machine learning expertise and computational resources for re-training from scratch or fine-tuning a network.

Test-time adaptation for groupwise registration (TAG)~\cite{ziyi2023template,he2024instantgroupinstanttemplategeneration} maps a group of images to a latent space using a VAE, computes an average of latent vectors, then decodes to estimate an atlas. 
While this rapidly produces atlases at inference, linearly averaging vectors in a VAE latent space
most often does not yield a representation that can be decoded into an unbiased deformable atlas. Further, this model must still be retrained for new imaging modalities or populations. In contrast, \method\: directly constructs group-specific atlases from warped images, ensuring fidelity to the data without distortions introduced by latent space aggregation. A single \method\: model can generate atlases for a wide variety of imaging modalities and population groups. 

\subpara{Flexible-size Inputs.} Recent methods have employed a variety of mechanisms that are flexible to input size, in other applications. For example, in-context learning methods use a flexible-sized input set of input-output example pairs to guide a new image-processing task at inference~\cite{butoi2023universeg,czolbe2023neuralizer}. Other methods use attention mechanisms across different inputs to aggregate information among volume slices~\cite{xu2022svort} or tabular data~\cite{kossen2021self}. While cross-attention and variants have been effective for many tasks in vision, they have quadratic memory complexity. At each iteration, our model requires a large set of $3$D volumes. Using cross-attention would lead to infeasible memory requirements. In contrast, we propose a flexible feature sharing mechanism with linear complexity to produce central atlases for large groups of $3$D images.


\section{Methods}

\subsection{Background}
Given two images $\vx_1,\vx_2 \subset \R^{d}$, deformable registration seeks a nonlinear mapping $\phi:\vx_1 \rightarrow \vx_2$ that warps one image into the space of the other. The deformation $\phi$ attempts to align the underlying anatomy captured by the images while maintaining a well-behaved map, and it is traditionally computed by optimization:
\begin{equation}
\label{eqn:reg-generic}
    \argmin_{\phi} \hspace{0.3em} \mathcal{L}_{sim}\left(\vx_2, \vx_1 \circ \phi \right) + \lambda \mathcal{L}_{reg}\left(\phi \right),
\end{equation}
where $\mathcal{L}_{sim}\left(\vx_2, \vx_1  \circ \phi \right)$ measures similarity between image $\vx_2$ and the warped image $\vx_1 \circ \phi$, $\mathcal{L}_{reg}(\cdot)$ regularizes the map $\phi$, and $\lambda$ is a hyperparameter that balances the two. 

Many population-based studies involve groupwise analyses. \emph{Group registration} aligns a collection of $m$ images $\mathcal{X}_m=\{\vx_i\}_{i=1}^m$ to an explicit image template $\vt$,
\begin{equation}
\label{eqn:group-reg}
    \argmin_{\phi_1,\ldots,\phi_m} \quad \sum_{i=1}^m \mathcal{L}_{sim}\left(\vt, \vx_i  \circ \phi_i \right) + \lambda \mathcal{L}_{reg}\left(\phi_i \right).
\end{equation}
In many scenarios, an explicit template is not available. One can be constructed by iterating a template estimation step, $\bar\vt = \frac 1 m \sum \vx_i \circ \phi_i$, and the groupwise optimization (\ref{eqn:group-reg}) until convergence. However, this is computationally expensive and does not scale well to large populations.

Machine learning approaches for pairwise registration use a neural network to predict $\phi$ as a function of the input images: $f_{\theta}(\vx_1,\vx_2)=\phi$, where $f$ is a neural network parameterized by $\theta$. Pairwise registration is rapidly computed by a single forward pass of a trained network. Recent methods~\cite{dalca2019learning,dey2021generative,ding2022aladdin,sinclair2022atlas} also estimate a common template $\vt$ together with parameters $\theta$ in a network, $f_{\theta}(\vt,\vx)=\phi$.

In this work, we develop a model to directly predict a group-specific set of deformation fields to a central template space. We formulate template construction as a group registration problem given a variable number of inputs.


\subsection{Flexible Group Registration}
Given a set $\mathcal X_m$ of $m$ images from a dataset $\mathcal X=\{\vx_i\}_{i=1}^n$, we seek to map the images to a space central to~$\mathcal{X}_m$. The set $\mathcal{X}_m$ could be an entire population or a subgroup of  patients representing an underlying demographic or condition.

Let $\phi_i: \R^d \rightarrow \R^d$ represent the map from $\vx_i$ to a central latent image space. We model a function $f_{\theta}(\{\vx_1,\ldots,\vx_m\}) = \{\phi_1,\ldots,\phi_m\}$ using a convolutional neural network with learnable parameters $\theta$. The number of parameters of $\theta$ is independent of the group size $m$.

To achieve desirable group registration, we construct $f$ to satisfy the following desiderata:
\begin{itemize}[leftmargin=20pt]
    \item Flexible to group size: $f$ takes as input a variable number of $m$ images, and computes $m$ maps $\phi$ to a group-specific central space.
    \item Fast: Computation of $\{\phi_i\}_{i=1}^m$ can be done in a single forward pass of an efficient network.
    \item Generalizable: generalize to unseen datasets $\mathcal Y_m$.
    \item Unbiased: the images in~$\mathcal{X}_m$ map to a space \textit{central} to that set: $\mathrm{mean}(\{\phi_i \}_{i=1}^m)=0$.
    \item Aligned: Images $\{\vx_i \circ \phi_i \}_{i=1}^m$ mapped to the template space are anatomically aligned.

\end{itemize}
\noindent Satisfying the desiderata leads to a model that can produce flexible templates for user-defined groups on demand. We introduce new methods to achieve these properties below.

\begin{figure*}[!t]
\centering
\includegraphics[width=\linewidth]{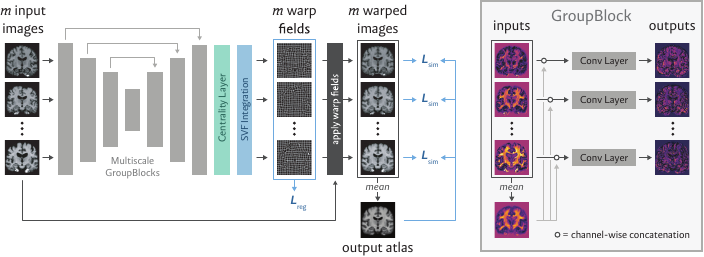}
\caption{\method\,  architecture diagram. The model takes in a variable group of $m$ images and constructs an atlas specific to that group. At each layer of the UNet, the proposed \texttt{GroupBlock} mechanism replaces standard convolution kernels. Specifically, it computes the elementwise mean of the intermediate  features across the group, and concatenates the resulting features  with the individual features. The mechanism enables group interaction by sharing summarized input features across the group.  
The network outputs $m$ velocity fields mapping images to the group-specific template space. A centrality layer removes any global bias in the average velocity field, before integration and warping the images. The output is a central template representing the shared anatomy of the input group.}
\label{fig:architecture}
\end{figure*}

\subsection{Model}
Figure~\ref{fig:architecture} gives an overview of the network architecture of \method. The network takes a group of a variable number of images and predicts diffeomorphic transformations to a central template space specific to the group. At each network layer, we share features across the inputs using the proposed \texttt{GroupBlock} layer. The network outputs stationary velocity fields, which are then adjusted by a centrality layer to produce an unbiased atlas. 

\subpara{Convolution Layer for Variable Group Size.}
\label{s:nonparam-layer}
We propose \texttt{GroupBlock}, a convolutional layer that combines image features across a group. As group registration seeks to align images to a central space, feature communication is helpful to produce a subgroup alignment. 

We use a summary statistic to aggregate group features, and communicate the statistic back to each individual group element. Let $c^{(l)}_i$ represent the feature map for input image $i$ at network layer $l$. The \texttt{GroupBlock} layer aggregates information as follows:
\begin{align*}
\bar{c}^{(l)} &= s (\{c_1^{(l)},\ldots,c_m^{(l)}\})  \\
c^{(l+1)}_i &= \mathrm{Conv}\left( \left [c_i^{(l)}  \big \| \bar{c}^{(l)} \right ] ; \theta^{(l)} \right),
\end{align*}
where $s(\cdot)$ is the summary statistic across the group dimension, $\left[\cdot \| \cdot \right]$ is the concatenation operation along the channel dimension, and $\mathrm{Conv}$ is a convolutional layer with parameters $\theta^{(l)}$. We use the mean as our summary statistic. 


\subpara{Network.}
We modify the popular UNet architecture~\cite{ronneberger2015u}, employing a multi-scale structure with residual connections. We replace the standard $\mathrm{Conv}$ layers with the proposed \texttt{GroupBlock} feature sharing layer (\S\ref{s:nonparam-layer}). The network takes as input a group of $m$ images and outputs $m$ $d$-dimensional stationary velocity fields (SVF). 

We use the standard SVF representation of a diffeomorphism~\cite{ashburner2007fast,dalca2019learning}. The deformation field is defined through the ordinary differential equation:
$\frac{\partial{\phi}_v^{(t)}}{\partial t}= v \circ \phi_v^{(t)}$, where $v$ is the velocity field. The field is defined for $t \in [0,1]$ with $\phi_v^{(0)}=Id$, the identity map. The deformation field is obtained by integrating $v$ using the scaling and squaring method~\cite{ashburner2007fast,dalca2019unsupervised}.


\subpara{Centrality Layer.}
Constructing an atlas central to the population group is key to performing unbiased downstream analyses, such as in quantifying anatomical shape differences without bias to any particular structure or subject. A central atlas is one that is ``close'' to the target population. Many learning approaches use a regularization term to minimize the mean displacement field~\cite{dalca2019learning,dey2021generative}. 

We construct a layer that produces a group-specific central template by construction. We subtract the groupwise mean from the output velocity fields: $v_i = v_i^{(L)} - \bar{v}^{(L)}$, where $v_i^{(L)}$ is the final output velocity field for image $i$, and $\bar v^{(L)}$ is the group mean. This centers the velocity fields in the zero-mean Lie subspace. 

\subpara{Template Construction.}
Given a trained network $f_{\theta}$, we can construct a template $\vt$ by aggregating the warped images of the group $\mathcal X_m$: $\vt = g \left( \vx_1 \circ \phi_1, \ldots, \vx_m \circ \phi_m \right)$. We use the \texttt{mean} operation for $g$.

To apply the map to the group of images, we integrate the SVF to obtain a diffeomorphic displacement field~\cite{ashburner2007fast,dalca2019unsupervised}. We then use a spatial transformation function~\cite{jaderberg2015spatial} to warp the images to the central space. The spatial transformer performs a pullback operation with linear interpolation.

\subsection{Auxiliary Structural Information} 

The use of anatomical labelmaps during training of learning-based registration often improves substructure alignment~\cite{balakrishnan2019voxelmorph}. 
When segmentation maps are available for some of the images in a set, we use this information to form an atlas segmentation map. Let $\mathrm{seg}[\vx_i]$ indicate the probabilistic segmentation labelmap of the $K$ structures for image $\vx_i$. We construct the labelmap of the template, $\mathrm{seg}[\vt]$, by taking the set-wise average of the warped probability maps $\mathrm{seg}[\vt] = \operatorname{mean}_m\{\mathrm{seg}[\vx_1] \circ \phi_1, \ldots, \mathrm{seg}[\vx_m] \circ \phi_m \}$. 

\subsection{Synthetic Training} To aid generalization to unseen modalities, we 
also train on images synthesized from brain tissue segmentations. For each synthetic training group, we sample $K$ random values uniformly corresponding to $K$ structures. We then use a domain randomization procedure~\cite{gopinath2024synthetic} to randomly sample intensity values for each structure, along with a variety of noise patterns and artifacts.  This yields groups of synthetic images, where each group exhibits random intensity distributions and tissue contrasts. Supplementary Fig.~\ref{fig:example-synth} presents a representative set of example synthetic images.

\subsection{Loss}
We maximize alignment between the images and anatomical structures of the group and the constructed template, while maintaining a smooth map. For a single image, the loss is computed as:
\begin{align}
\label{eqn:loss-ovverall}
    \mathcal{L} \left ( \phi_i   \right) = &\mathcal{L}_{sim} \left(\vt, \vx_i \circ \phi_i  \right) + \lambda \mathcal L_{reg} \left( \phi_i \right) \nonumber \\
    & + \gamma \mathcal L_{struc} \left(\mathrm{seg} \left[\vt \right], \mathrm{seg} \left[\vx_i \right] \circ \phi_i \right).
\end{align}
The first term $\mathcal L_{sim}$ measures pairwise similarity between image $\vx_i$ and the template $\vt$. We use the normalized cross-correlation objective. The second term regularizes the deformation field to be smooth, $\mathcal L_{reg}(\phi_i) = \| \nabla \phi_i \|^2$.  When label maps are available during training, we use the third (auxiliary) loss term to align the structures of the training set with the constructed template, using soft-Dice.

Our complete  group loss is $\mathcal{L}(\phi_1,\ldots,\phi_m) = \frac{1}{m}\sum_{\{i : \vx_i \in \mathcal X_m\}}\mathcal{L}(\phi_i)$. Since the template $\vt$ is constructed by averaging warped images of the group, the loss is dependent on all images of the group.

\section{Experiments}

We evaluate \method{} using 3D brain MRI brain scans, a common setting for atlas construction. We compare \method{} against iterative and learning-based approaches in terms of speed, centrality, and accuracy. We also test whether \method{} generalizes to new datasets, imaging modalities, and populations that are unseen during training.

\subsection{Experimental Setup}
 
\paragraph*{Data.} We use four public 3D brain MRI datasets. Three datasets — OASIS-1, OASIS-3, and MBB — are used for training, validation, and testing, while IXI serves as an unseen test set.
OASIS-1~\cite{marcus2007open} includes T1-weighted (T1-w) scans of 416 subjects aged 18-96. A hundred OASIS-1 subjects of ages 60 years and older were diagnosed with mild to moderate Alzheimer's disease (AD), which is correlated with brain atrophy.
OASIS-3~\cite{lamontagne2019oasis} contains T1-w and T2-w MRI scans of subjects aged 42-95 years old. We use a subset of $1043$ subjects, with $210$ diagnosed with mild to severe cognitive dementia.
The Mind Brain Body dataset~\cite{babayan2018mind} includes T2-w and T2-FLAIR scans of 226 healthy subjects. For each training dataset, we randomly hold out 20\% of the subjects for testing, and split the rest into 85\% for training and 15\% for validation. Each split includes an equal mix of healthy and abnormal subjects of all age ranges. We use the same model for all experiments.

Lastly, to evaluate generalization, we hold out the IXI dataset~\cite{ixi_dataset}. We arbitrarily select the Guys Hospital site within IXI and retrieve T1-w, T2-w, and PD-w MRI scans of 319 adult subjects.  Importantly, the PD-w MRI modality is not included in any of the training datasets used by our model. These datasets span a large age range and include a mix of disease states and imaging modalities, simulating real-world population studies.

\subpara{Implementation details.} During training, all images within a sampled group have the same acquisition modality. We apply augmentations, including random exponential scaling, intensity inhomogeneities, and per-voxel Gaussian noise. Additionally, 50\% of the sampled training groups contain synthetic images instead of real acquisitions. For preprocessing, using ANTs~\cite{tustison2021antsx}, we affinely align each 3D scan to a common 1-mm isotropic affine reference used in~\cite{hoffmann2021synthmorph,hoffmann2023anatomy}. We extract brain tissue signal using SynthStrip~\cite{hoopes2022synthstrip} and generate segmentation maps of 30 unique anatomical brain structures using SynthSeg~\cite{billot2023synthseg}.

\begin{table*}[t]
\footnotesize
  \caption{Atlas construction evaluation on $319$ brain volumes from IXI. While all baselines were trained or optimized on the full dataset, \method{} was not, demonstrating its ability to generalize to entirely new datasets. \method{} also generalizes to the PD-w modality not seen during training, demonstrating its capabilities on unseen imaging modalities. $^*$ indicates statistical significance ($p<0.01$).}
  \label{tab:results-all}
  \label{tab:atlas-results}
  \centering
  \begin{tabular}{clcccc}
    \toprule
    \textbf{Modality} & \textbf{Method} &  \textbf{Construction} \textbf{time} (min.) ($\downarrow$) &\textbf{Dice} ($\uparrow$) & \textbf{Folds} ($\downarrow$)   &  \textbf{Centrality} $\times 10^{-2}$ ($\downarrow$)  \\ 
    \midrule
    
     \multirow{3}{*}{\makecell{\\  T1-w}} &  
ANTs \cite{avants2010optimal} & 4345.20 & $0.863\pm0.075$ & $524.2\pm580.04$ &  $10.4\pm30.67$ \\
&AtlasMorph \cite{dalca2019learning} & 1141.50 & $0.894\pm0.015$ & $47.9\pm29.22$ &   $7.8\pm19.09$ \\
&Aladdin \cite{ding2022aladdin} & 325.20 & $0.885\pm0.01$ & $\mathbf{0.0\pm0.0^*}$  & $106.8\pm97.6$ \\
&Ours & \textbf{10.50} & $\mathbf{0.913\pm0.006^*}$ & $1.1\pm1.55$ &  $\mathbf{1.4\pm4.32^*}$ \\
      
      \midrule
      
      \multirow{3}{*}{\makecell{\\ T2-w}} & 
ANTs \cite{avants2010optimal} & 4380.60 & $0.862\pm0.071$ & $522.6\pm476.86$ &  $18.6\pm44.286$ \\
&AtlasMorph \cite{dalca2019learning} & 831.60 & $0.882\pm0.018$ & $57.5\pm31.935$ &  $7.8\pm19.34$ \\
&Aladdin \cite{ding2022aladdin} & 261.00 & $0.875\pm0.012$ & $\mathbf{0.0\pm0.125^*}$ &  $771.5\pm744.309$ \\
&Ours & \textbf{10.40} & $\mathbf{0.906\pm0.007^*}$ & $2.0\pm2.49$ &  $\mathbf{1.5\pm4.683^*}$ \\
      \midrule
      
\multirow{3}{*}{{\makecell{\\ PD-w }}} & 
ANTs \cite{avants2010optimal} & 4320.20 & $0.856\pm0.069$ & $313.1\pm359.9$ &  $12.4\pm32.805$ \\
&AtlasMorph \cite{dalca2019learning} & 959.00 & $0.884\pm0.018$ & $40.5\pm26.10$ &  $7.4\pm19.483$ \\
&Aladdin \cite{ding2022aladdin} & 163.80 & $0.849\pm0.029$ & $\mathbf{0.0\pm0.0^*}$ &  $1175.7\pm1731.773$ \\
&Ours & \textbf{7.80} & $\mathbf{0.900\pm0.009^*}$ & $1.601\pm0.205$ &  $\mathbf{0.9\pm3.02^*}$ \\

    \bottomrule
  \end{tabular}
\end{table*}

%

\looseness=-1
We train using the Adam optimizer~\cite{kingma2014adam} with a learning rate of $10^{-4}$. The field regularization hyperparameter is set to $\lambda=1.0$ and the segmentation-loss weight is $\gamma=0.5$, both chosen via grid search (Suppl. Sec.~\ref{s:sup-sensitivity}). At each training iteration, we randomly sample $m=[2,12]$ images to form a group and train for $80,000$ iterations, using the final saved model. All models are trained on a single RTX8000 GPU. The ANTs experiments and all runtime evaluation results were done on an Intel(R) Xeon(R) Gold 5218 CPU. 

\looseness=-1
\subpara{Baselines.} We evaluate SyGN~\cite{avants2010optimal}, a widely-used iterative atlas construction method from the ANTs library~\cite{tustison2021antsx}. Additionally, we compare against AtlasMorph, a learning-based atlas constructor~\cite{dalca2019learning} that explicitly \textit{learns} an atlas to best fit the training data. 
For AtlasMorph, we set the deformation field regularization hyperparmeter to $\lambda=0.1$, as determined via cross-validation. Both \method{} and AtlasMorph use the same core registration network. 

We also evaluate Aladdin~\cite{ding2022aladdin}, a learning-based method that constructs an average reference atlas during training by learning pairwise registrations. At test time, this atlas serves as the registration target, enabling the generation of new atlases for different population groups. Since Aladdin constructs modality-specific atlases, we train a separate model (with the same capacity as our network) for each modality in our dataset using an optimal regularization loss weight of 10,000, a similarity loss weight of 10, and an image pair loss weight of 0.2, all determined using a grid search. Both AtlasMorph and Aladdin models are trained for 50,000 iterations, followed by 1,500 finetuning iterations per population subgroup to estimate a group atlas at test-time. 

\begin{table}[b]
\caption{Sub-group atlas construction results. Reported scores are averaged across atlases constructed using subgroups of [$5,10,20,\ldots,60$]. $^*$ indicates statistical significance ($p<0.01$).}
\label{tab:results-subgroup}
\scriptsize
\setlength{\tabcolsep}{3.25pt}
\centering
\begin{tabular}{lcccc}
\toprule
\textbf{Method} &  \makecell{\textbf{Run time} \\ \ (min.) ($\downarrow$)} &\makecell{\textbf{Dice} \\ \textbf{Transfer}} ($\uparrow$) & \textbf{Folds} ($\downarrow$)   &   \makecell{\textbf{Centrality}\\ $\times 10^{-2}$ ($\downarrow$)} \\
\midrule
ANTs~\cite{avants2010optimal} & $436\pm0.4$ & $0.875\pm0.009$ & $447\pm110$ & $8.7\pm0.1$ \\
AtlasMorph~\cite{dalca2019learning} & $17\pm1.4$ & $0.893\pm0.005$ & $50.0\pm8.7$ & $9.7\pm0.1$ \\
Aladdin~\cite{ding2022aladdin} & $12\pm0.1$ & $0.877\pm0.004$ & $\mathbf{0.0\pm0.0^*}$ & $173\pm3.7$ \\
Ours & $\mathbf{1.5\pm0.0}$ & $\mathbf{0.904\pm0.002^*}$ & $1.3\pm0.4$ & $\mathbf{1.4\pm0.04^*}$ \\
\bottomrule
\end{tabular}
\end{table}

\subpara{Evaluation.}
We assess the effectiveness of atlas construction techniques in rapidly generating central atlases for new populations. To evaluate registration quality, we compute the Dice score to assess how well the atlas aligns with warped subject scans. We assess field regularity and topology by computing the determinant of the Jacobian of the map, $\det J_{\phi}(p) =\det \left (\nabla \phi \left (p \right) \right)$ at each voxel $p$. Locations where $\det J_{\phi}(p) < 0$ represent folded regions breaking local injectivity. Additionally, we measure atlas centrality by reporting the mean displacement field $\|\bar{\vu}\|^2$. Statistical significance is determined using a paired t-test with $p<0.01$. 

\subpara{Segmentation transfer.} As atlases are commonly used for segmentation by warping atlas labels to new target images, we evaluate each method's segmentation performance.  Each atlas is estimated using half the subgroup ($\frac m 2$ images). We randomly sample $\frac m 2$ segmentation label maps to generate the atlas segmentation mask, which is then transferred to the remaining $\frac m 2$ images. Segmentation quality is assessed using the Dice score.

\subsection{Results}

\subsubsection{Generalizing to Unseen Datasets and Modalities} 
\label{s:unseen-results}

%
Table~\ref{tab:results-all} presents results for all methods on the IXI dataset, which was entirely held-out for \method{}'s training and validation. \method{} produces atlases over a $100\times$ faster than ANTs and AtlasMorph, and $30\times$ faster than Aladdin. It consistently achieves the highest Dice score, indicating better anatomical alignment even when constructed on unseen data at test time in a single forward pass. Additionally, \method{} yields regular deformation fields with negligible folding and significantly lower bias in the displacement fields, indicating that the constructed atlases are central.  

\begin{figure}[b]
    \centering
    \includegraphics[width=\linewidth]{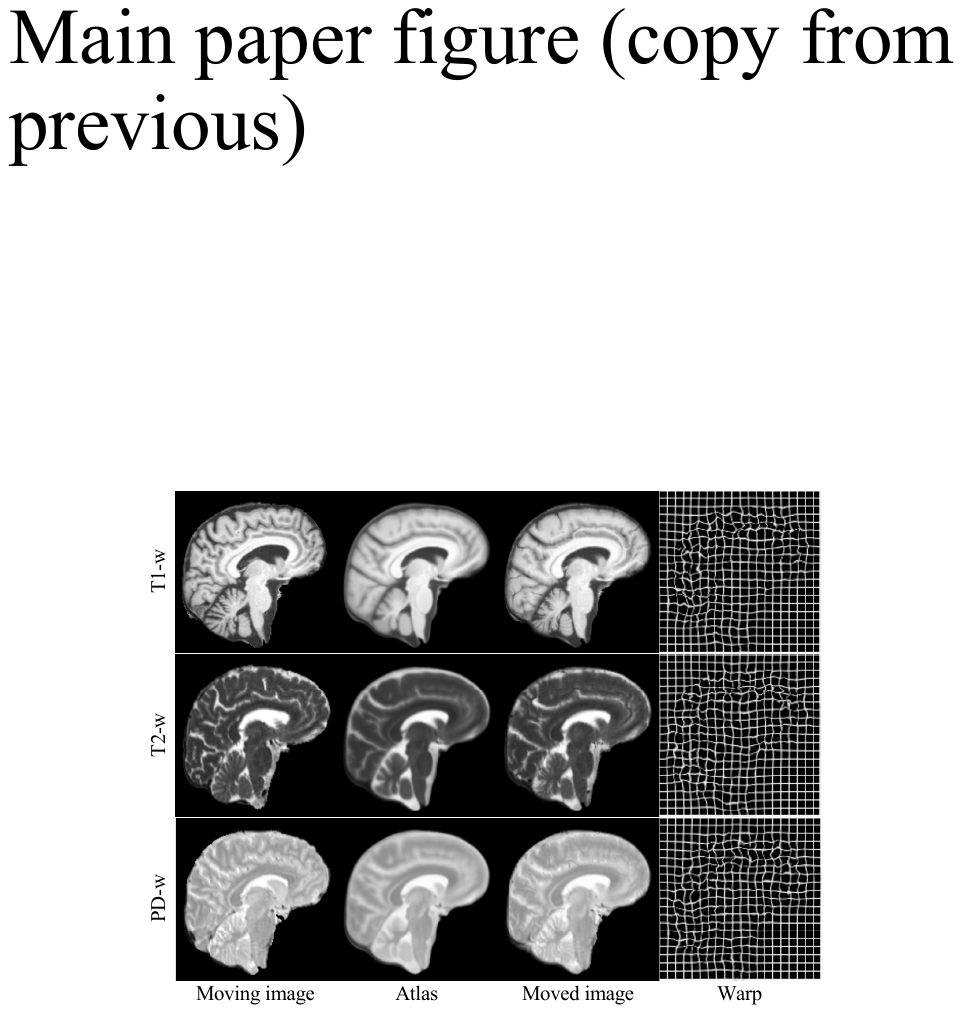}
    \caption{Example images and warps to the atlas constructed using the IXI dataset, for three subjects and three modalities.}
    \label{fig:example-warps-main}
\end{figure}

\begin{figure*}[t]
     \centering
     \includegraphics[width=0.975\linewidth]{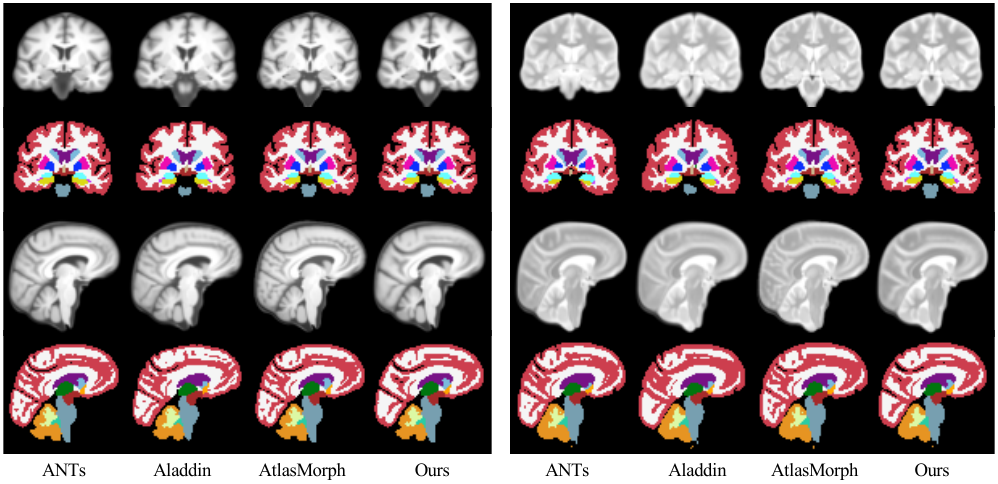}
     \caption{Atlases constructed on the IXI T1-w (left) and IXI PD-w (right) image modality. All baseline methods used the dataset for training or optimization, while our method was not trained on the IXI data. Further, our method was never trained on PD-w images, yet generalizes to this modality. }
     \label{fig:whole-atlas-figure}
 \end{figure*}

Fig.~\ref{fig:example-warps-main} visualizes sample registration predictions for each modality in IXI and Fig.~\ref{fig:whole-atlas-figure} illustrates example atlases for IXI T1-w and PD-w. Despite never having been trained on this dataset nor having seen the PD-w imaging modality during training, \method{} estimates atlases that yield high group alignment in only minutes, demonstrating its potential for scientific studies requiring specific atlases. 
We provide additional examples in Supplemental Fig.~\ref{fig:images-and-warps}.

\subsubsection{Standard Atlas Construction}
\label{s:oasis3-atlas-results}
We now evaluate the ability of \method{} to construct population atlases across different age groups and disease states. Specifically, we construct an atlas on the OASIS-3 T1-w test dataset. All baseline models were trained and validated on the test set. Table~\ref{tab:results-OASIS-3} shows that \method{} achieves the highest Dice score while producing atlases $30-400$ times faster than the baseline methods. 

\begin{figure}[b]
  \centering
    \includegraphics[width=0.4\textwidth]{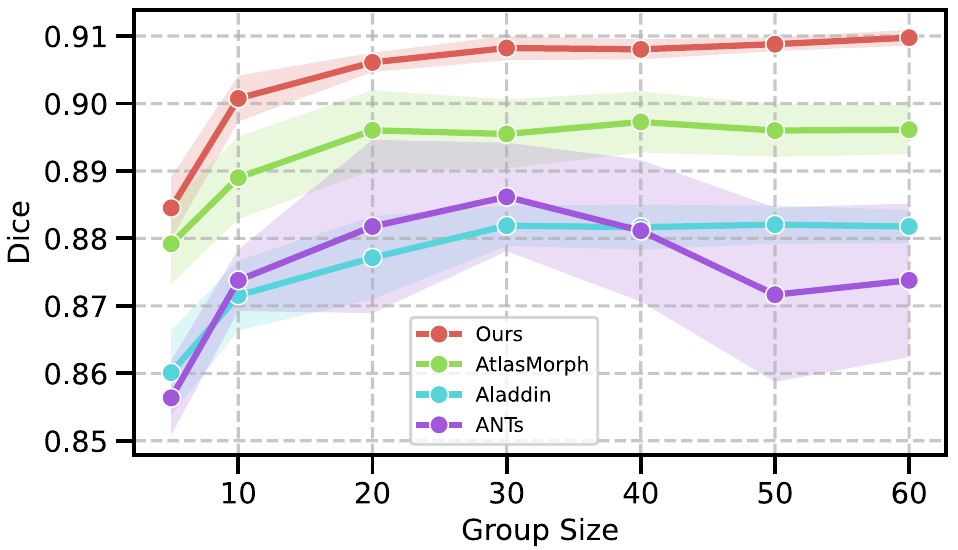}
  \caption{Segmentation transfer performance when varying the number of images used to construct an atlas. Data is taken from the IXI T1-w dataset, which our model did not have access to during training. Our method consistently outperforms the baselines.}
  \label{fig:subgrp-plot}
\end{figure}

\subsubsection{Subgroup Atlas Construction}
\label{sec:subgroup}

\method{} enables the rapid construction of subgroup atlases for granular population analyses. We evaluate atlases conditioned on age, age and disease state, as well as random subgroupings of the population.

\begin{figure*}[hbt!]
    \centering
    \includegraphics[width=\linewidth]{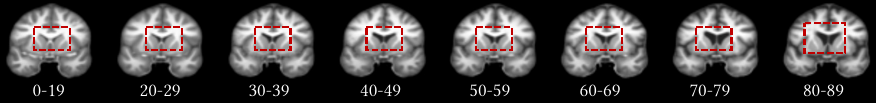}
    \caption{Atlases conditioned on age for healthy subjects in OASIS-1. Ventricle enlargement (red boxes) is observed across time, consistent with neurodegeneration with aging.}
    \label{fig:OASIS-1-conditional-healthy}
\end{figure*}

\subpara{Random Subgroup Analysis.} We quantify the effect of subgroup size on atlas quality using the held-out IXI T1-w dataset. Subgroups of $[5,10,20,\ldots,60]$ images are randomly sampled, with half used to construct the atlas segmentation and the other half used for evaluation. As in  Section~\ref{s:unseen-results}, the baselines were trained or optimized on this dataset, whereas \method{} was not exposed to any IXI T1-w data during training or validation.

\begin{table}[b]
\caption{Atlas estimation results on 212 subjects from the OASIS-3 T1-w test set. 
$^*$ indicates statistical significance ($p<0.01$).
}
\label{tab:results-OASIS-3}
\scriptsize
\setlength{\tabcolsep}{3.25pt}
\begin{tabular}{lcccc}
\toprule
\textbf{Method} &  \makecell{\textbf{Run time} \\ \ (min.) ($\downarrow$)} &\makecell{\textbf{Dice}\\\textbf{Transfer}} ($\uparrow$) & \textbf{Folds} ($\downarrow$)   &   \makecell{\textbf{Centrality}\\ $\times 10^{-2}$ ($\downarrow$)} \\
\midrule
ANTs~\cite{avants2010optimal} & 2858 & $0.886\pm0.017$ & $765\pm877$ & $9.5\pm25.6$ \\
AtlasMorph~\cite{dalca2019learning} & 688 & $0.881\pm0.024$ & $50.2\pm31.9$ & $8.0\pm0.2$ \\
Aladdin~\cite{ding2022aladdin} & 277 &  $0.878\pm0.016$ & $\mathbf{0.0\pm0.07}^*$ & $175.9\pm1.8$ \\
Ours & \textbf{5.9} & $\mathbf{0.910\pm0.014^*}$ & $1.2\pm2.3$ & $\mathbf{1.5\pm0.05}^*$ \\
\bottomrule
\end{tabular}
\end{table}

Fig.~\ref{fig:subgrp-plot} shows that \method{} consistently outperforms baselines, with performance improving as the subgroup size increases. Table~\ref{tab:results-subgroup} reports mean performance across subgroups, with \method{} showing better segmentation transfer while maintaining well-behaved deformation fields. Importantly, \method{} only requires $1.5$ minutes of inference time on a CPU, whereas baselines require fine-tuning or re-optimization, which is both time consuming and requires tens or hundreds of minutes.

\subpara{Age.} We first demonstrate \method{}'s ability to create appropriate atlases for user-defined subgroups by grouping healthy OASIS-1 subjects into age bins. We take normal subjects in the validation and test set, and bin them into age ranges $[0-19,20-29,\ldots,80-89]$. Fig.~\ref{fig:OASIS-1-conditional-healthy} presents qualitative results, showing anatomical changes consistent with normal aging, such as ventricular enlargement due to brain atrophy~\cite{apostolova2012hippocampal}. All atlases were generated in under a minute without any fine-tuning.

\begin{figure}[b]
    \centering
    \includegraphics[width=0.98\linewidth]{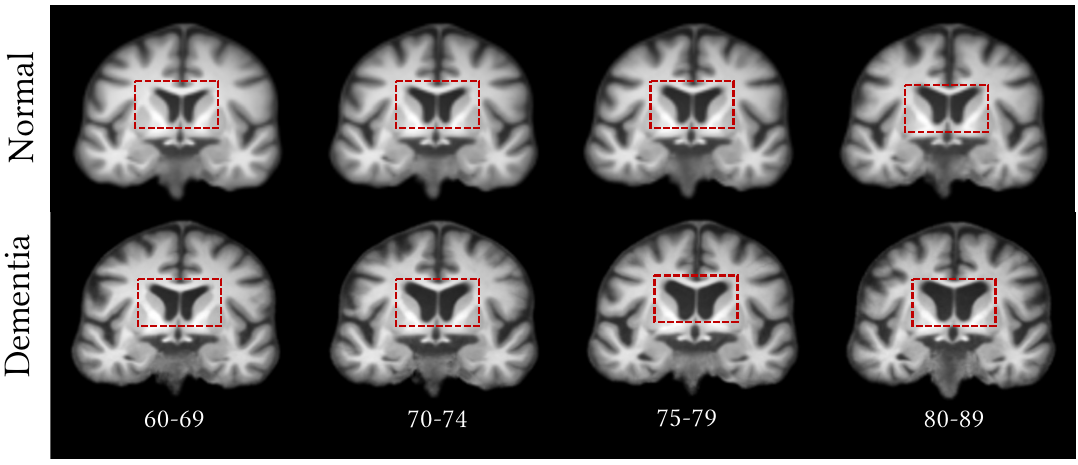}
    \caption{Atlases conditioned on age for normal subjects (top) and subjects with dementia (bottom) from OASIS-3. Visual differences indicate considerable enlargement of ventricles (red boxes) and atrophy of white matter when compared to normal subjects. }
    \label{fig:OASIS-3-T1-w-condition}
\end{figure}

\subpara{Diagnosis.} Lastly, we examine the effect of dementia on brain aging in the OASIS-3 (T1-w) dataset. We construct age-conditioned atlases separately for normal and dementia-diagnosed subjects. Fig.~\ref{fig:OASIS-3-T1-w-condition} compares brain atrophy across matched age groups. We observe substantial enlargement of the ventricles (outlined in red boxes) and deterioration of the white matter in the dementia group as compared to the controls, consistent with the literature~\cite{apostolova2012hippocampal,zhang2009white,nestor2008ventricular,kempton2011comprehensive}.  

\subsection{Ablation studies}
\label{s:ablation}
We quantify the effect of several key model components, including the centrality layer (CL), the \texttt{Group Block} (GB) mechanism with varying summary statistics (mean, variance, max), and training without the Dice Loss. Using the OASIS-1 dataset~\cite{marcus2007open}, we train our model for $50,000$ iterations and assess performance on the test set. 

Table~\ref{tab:ablation-oasis1} summarizes the results. The CL significantly reduced the centrality measure by $1000\times$, enabling unbiased atlas construction, although it led to a $1$ point decrease in Dice. The GB mechanism improved Dice by $1.4$ points with negligible degradation of field regularity. We observe no significant performance variation across the various summary statistics tested. Finally, the Dice loss improved performance by over $2$ Dice points. Taken holistically, each component strongly contributed to the \method{} performance.
We further quantify the impact on subgroup atlas construction in Supplemental Section~\ref{s:ablation-subgroup} and observe similar trends. Additionally, we assess the impact of training with synthetic data in Supplemental Section~\ref{s:ablation-synth}, which improved IXI dataset performance by up to $1.8$ Dice points while maintaining field regularity, demonstrating better generalization.

\begin{table}[b]
\caption{Model ablations on the Centrality Layer, Group Block mechanism, and Dice loss on the OASIS-1 test set. All proposed components improved atlas construction performance.}
\label{tab:ablation-oasis1}
\scriptsize
\centering
\begin{tabular}{lcccc}
\toprule
\textbf{Ablation} &  \makecell{\textbf{Dice} \\ \textbf{Transfer} ($\uparrow$)} & \textbf{Folds} ($\downarrow$)   &   \makecell{\textbf{Centrality}\\ $\times 10^{-3}$ ($\downarrow$)} \\
\midrule
no CL, GB(mean) & $0.892\pm0.018$ & $\mathbf{0.0\pm0.0}$ & $16125\pm11494$ \\ 
CL, no GB  & $0.870\pm0.021$ & $0.1\pm0.3$ & $\mathbf{9.9\pm27.4}$ \\
CL, GB(var)  & $0.883\pm0.020$ & $1.5\pm2.8$ & $12.8\pm59.27$ \\
CL, GB(max)  & $0.880\pm0.019$ & $1.5\pm2.7$ & $12.6\pm46.69$ \\
CL, GB(mean)  & $0.884\pm0.020$ & $1.1\pm1.9$ & $12.0\pm39.48$  \\
CL, GB(mean), Dice & $\mathbf{0.919\pm0.011}$ & $5.4\pm7.5$ & $18.6\pm61.31$ \\
\bottomrule
\end{tabular}
\end{table}




 
  
\section{Discussion}

\noindent\textbf{Limitations and future work.} \method{} has several avenues for extensions. For example, as it assumes diffeomorphic transformations, \method{} cannot accurately construct atlases for neuroimages with topology-changing pathologies. However, this can be addressed by using pathology masks when calculating losses in training~\cite{brett2001spatial}. Additionally, \method{} is currently only trained for neuroimages, but can be trained on anatomy-\textit{agnostic} synthetic data~\cite{hoffmann2021synthmorph,dey2024learning} to estimate atlases for arbitrary applications. 
Lastly, our implementation stores all activations in memory at inference, potentially limiting higher group sizes with large 3D volumes in memory-constrained settings. 

\subpara{Conclusion.} We presented \method, a test-time atlas construction framework that works with unseen imaging modalities and any number of input images--without retraining. At its core, \method{} leverages a novel convolutional layer for \textit{groups} of images, independent of the number of input samples, enabling efficient and scalable atlas generation.
\method{} produces unbiased atlases for arbitrary inputs with comparable (and often better) performance, while also being over 100 times faster than previous approaches that require either solving an optimization problem or retraining a model. 
By making high-quality atlas construction fast, accessible, and adaptable, \method{} potentially unlocks new avenues for biomedical research, enabling computational anatomy studies that were previously impractical due to computational constraints.

\section*{Acknowledgements}
Marianne Rakic was added as an author for contributions made since the CVPR deadline.

 We thank Zack Berger for help in proofreading. Support for this research was provided in part by Quanta Computer Inc. project AIR, the NIH BICCN grants U01 MH117023 and UM1 MH130981, NIH BRAIN CONNECTS U01 NS132181, UM1 NS132358, NIH NIBIB R01 EB023281,
R21 EB018907, R01 EB019956, P41 EB030006, NIH NIA R21 AG082082, R01 AG064027, R01 AG016495, R01 AG070988, the NIH NIMH UM1 MH130981, R01 MH123195, R01 MH121885, RF1 MH123195, NIH NINDS
U24 NS135561, R01 NS070963, R01 NS083534, R01 NS105820, R25 NS125599, NIH NICHD R00 HD101553, NIH R01 EB033773,
and was made possible by the resources provided by NIH Shared
Instrumentation Grants S10 RR023401, S10 RR019307, and
S10 RR023043. Additional support was provided by the NIH Blueprint for Neuroscience Research U01 MH093765, part of the multi-institutional Human Connectome Project. Much of the computation resources was performed on hardware provided by the Massachusetts Life Sciences Center.

{
    \small
    \bibliographystyle{ieeenat_fullname}
    \bibliography{biblio}
}

\clearpage
\setcounter{page}{1}
\maketitlesupplementary

\section{Ablation Studies}
\label{s:supp-ablation}
We conduct several ablations to quantify the effect of individual components of the proposed model.

\begin{table}[b]
\caption{Model subgroup ablations. We aggregate performance on atlases created from random subgroups of [5,10,20,30,40] images from the OASIS-1 test set. The GB effectively shares group features, improving subgroup atlas construction.}
\label{tab:results-subgroup-ablations}
\scriptsize
\centering
\begin{tabular}{ccccc}
\toprule
\textbf{Ablation} &  \makecell{\textbf{Dice}} ($\uparrow$) & \textbf{Folds} ($\downarrow$)   &   \makecell{\textbf{Centrality}\\ $\times 10^{-3}$ ($\downarrow$)} \\
\midrule
GB (mean)+Dice  & $\mathbf{0.911\pm0.002}$ & $7.1\pm1.4$ & $18.7\pm0.5$ \\
GB (mean) & $0.879\pm0.005$ & $0.7\pm0.4$ & $13.8\pm1.4$ \\
GB (max) &  $0.878\pm0.005$ & $0.8\pm0.4$ & $14.3\pm1.2$ \\
GB (var) &  $0.878\pm0.006$ & $0.7\pm0.3$ & $14.0\pm1.3$ \\
no GB &  $0.862\pm0.006$ & $\mathbf{0.0\pm0.0}$ & $\mathbf{12.4\pm2.5}$ \\
\bottomrule
\end{tabular}
\end{table}

\subsection{Effect of Synthetic Data}
\label{s:ablation-synth}
We first evaluate the effect of training with and without synthetic data. We presents results on the generalization experiment of Section~\ref{s:unseen-results}. We evaluate on the held-out IXI dataset, quantifying the results on T1-w, T2-w, and PD-w image modalities. Table~\ref{tab:ablation-synth-ixi} presents the results. In all cases, the inclusion of synthetic data improves the segmentation transfer performance with negligible increase in centrality and number of folds.   

\begin{table*}[]
\centering
\caption{IXI held out dataset atlas construction results, comparing our method trained with and without synthetic data.}
\label{tab:ablation-synth-ixi}
\begin{tabular}{cccccc}
\toprule
\textbf{Modality}& \textbf{Method} &  \makecell{\textbf{Dice} \\ \textbf{Transfer}} ($\uparrow$) & \textbf{Folds} ($\downarrow$)   &  \textbf{Norm Disp.} ($\downarrow$) & \makecell{\textbf{Centrality}\\ $\times 10^{-3}$ ($\downarrow$)} \\
\midrule
T1-w & Ours (w/ Synth) &  $\mathbf{0.911\pm0.007}$ & $1.1\pm1.634$ & $1.659\pm0.204$ & $13.5\pm40.914$ \\
& Ours (no Synth)  & $0.894\pm0.011$ & $0.5\pm1.057$ & $1.552\pm0.171$ & $10.0\pm29.452$ \\
\midrule
T2-w &Ours (w/ Synth) & $0.904\pm0.008$ & $1.7\pm2.346$ & $1.74\pm0.209$ & $13.7\pm40.101$ \\
&Ours (no Synth) &  $0.888\pm0.013$ & $0.7\pm1.295$ & $1.611\pm0.181$ & $8.9\pm24.201$ \\
\midrule
PD-w & Ours (w/ Synth) &  $0.897\pm0.011$ & $0.6\pm1.299$ & $1.599\pm0.205$ & $8.9\pm27.473$ \\
& Ours (no Synth) &  $0.882\pm0.015$ & $0.3\pm1.176$ & $1.491\pm0.172$ & $6.5\pm19.39$ \\
\bottomrule
\end{tabular}
\end{table*}

\subsection{Model ablations}
\label{s:ablation-model}
We quantify the effect of several key model components on the OASIS-1 dataset, as described in Section~\ref{s:ablation}. Here, we assess the effect on subgroup atlas construction.

\begin{figure}[h]
    \centering
    \includegraphics[width=\linewidth]{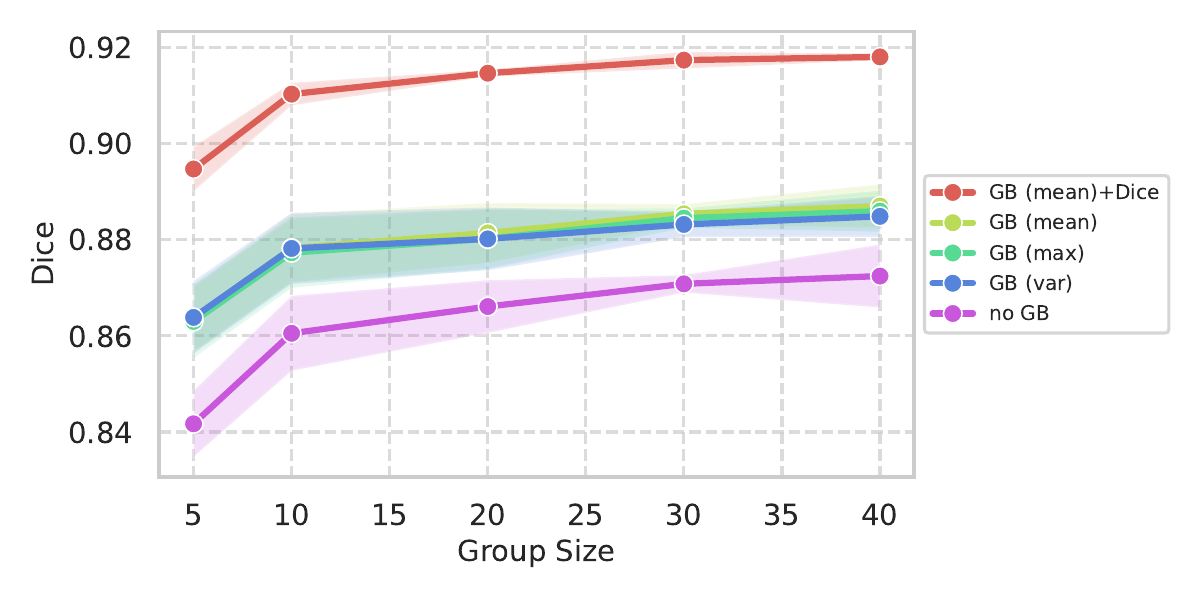}
    \caption{Subgroup atlas construction results across ablation studies on the GroupBlock mechanism. Shaded regions denote the 95\% confidence interval. Including the GB mechanism led to significant improvements in segmentation transfer compared to without. Further, training with the Dice loss led to a consistent improvement of up to $2$ Dice points. }
    \label{fig:subgroup-ablation}
\end{figure}

\subpara{Subgroup Atlas Construction}.
\label{s:ablation-subgroup}
We hypothesize that constructing atlases for homogeneous groups benefits more from within-group feature interactions than heterogeneous groups, by capturing set-specific information. To test this hypothesis, we split the OASIS-1 test set into random subgroups of $[5,10,20,30,40]$ images and quantify performance. Figure~\ref{fig:subgroup-ablation} presents the results on the segmentation transfer task.   Table~\ref{tab:results-subgroup-ablations} presents average results across all subgroups. The effect of the GroupBlock mechanism is immediately apparent, leading to a large increase in Dice score while maintaining well-behaved deformation fields. The improvement enabled by the Group Block mechanism is especially evident in homogeneous groups. For narrow atlas construction tasks, feature sharing within an image group is helpful to produce meaningful, group-specific atlases.


\section{Sensitivity Analysis}
\label{s:sup-sensitivity}
We quantify the sensitivity of our model performance to hyperparameters. Using the OASIS-1 validation set, we measure the effect of changing the regularization hyperparameter $\lambda$ and the Dice loss hyperparameter $\gamma$ in the produced atlas. Specifically, we measure the effect on Dice transfer, number of folds, and Centrality.

\begin{figure*}[]
    \centering
    \begin{tabular}{ccc}
    \includegraphics[width=.33\linewidth]{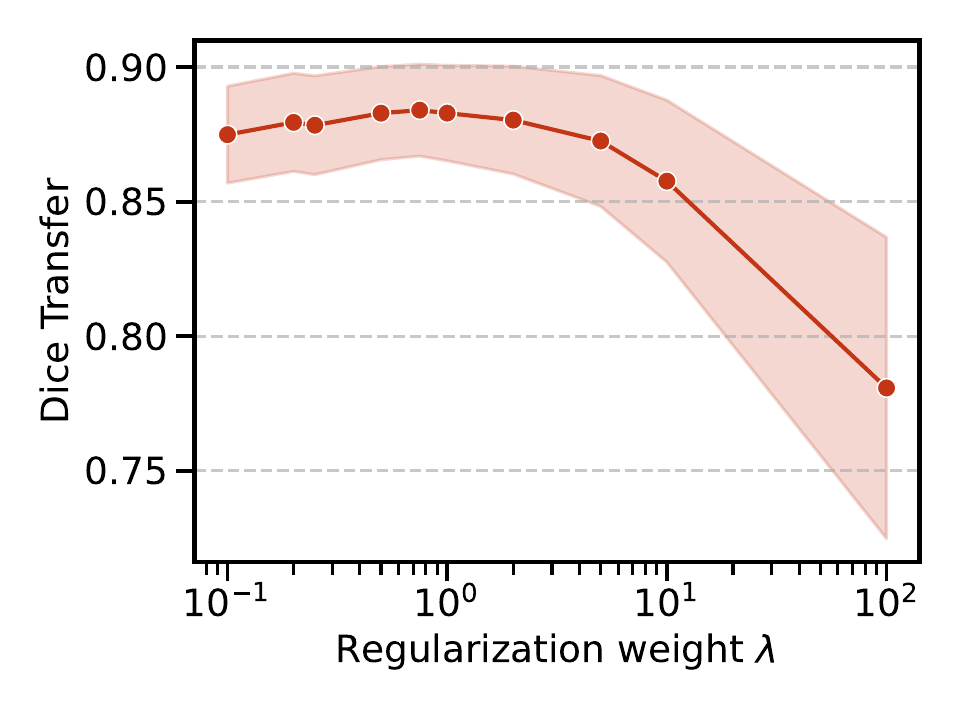} &
    \includegraphics[width=.33\linewidth]{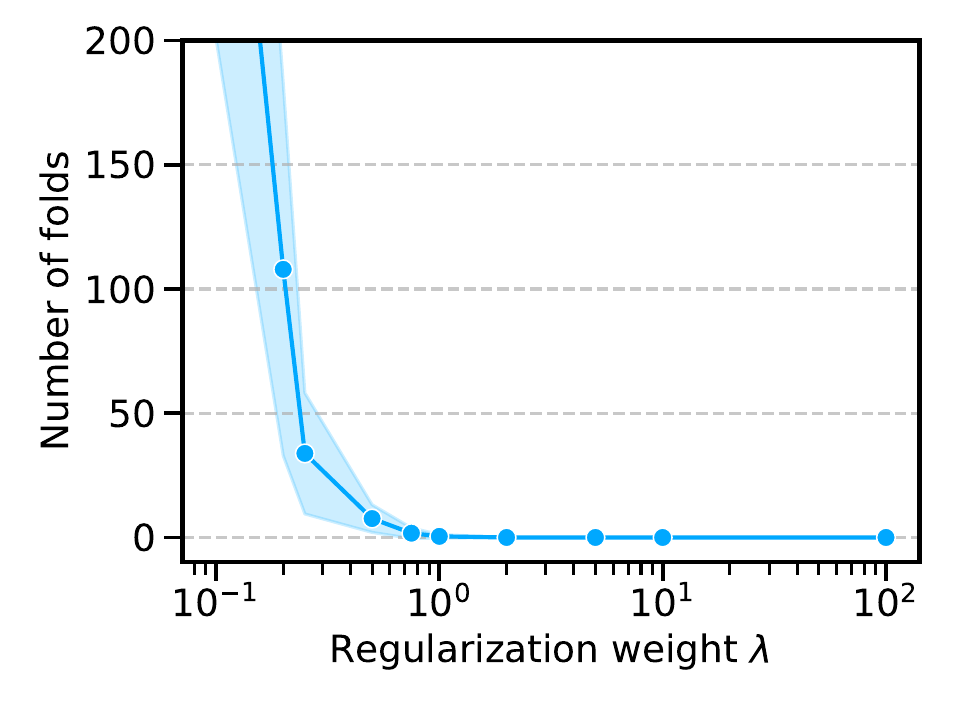} &
    \includegraphics[width=.33\linewidth]{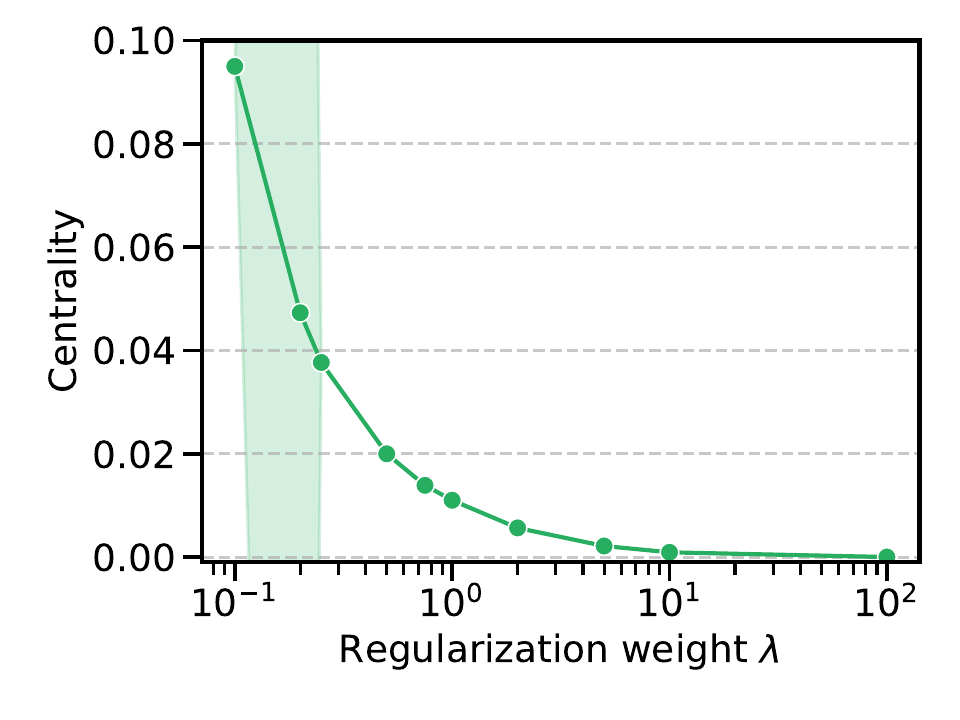}
    \\
    \end{tabular}
    \caption{Hyperparameter sweep over regularization weight $\lambda$ with Dice loss weight $\gamma=0$ on the OASIS-1 validation set. Shaded regions represent one standard deviation from the mean. Plots show the effect on Dice segmentation transfer, number of folded voxels, and Centrality. Our model shows consistent performance for $\lambda\in[0.5,\ldots,2]$, indicating robustness. We select $\lambda=1$ as it achieves a reasonable tradeoff between segmentation alignment and field regularity.}
    \label{fig:paramsweep-lambda}
\end{figure*}

Figure~\ref{fig:paramsweep-lambda} shows results while varying $\lambda$ and setting $\gamma=0$. We observe well behaved deformation fields with strong structural alignment for $\lambda \in [0.5,\ldots,2]$, indicating our model is robust to the choice of this hyperparameter. We set $\lambda=1$ for all experiments as it achieves a good tradeoff between structural alignment and smooth deformation fields.

\begin{figure*}[b]
    \centering
    \begin{tabular}{ccc}
    \includegraphics[width=.33\linewidth]{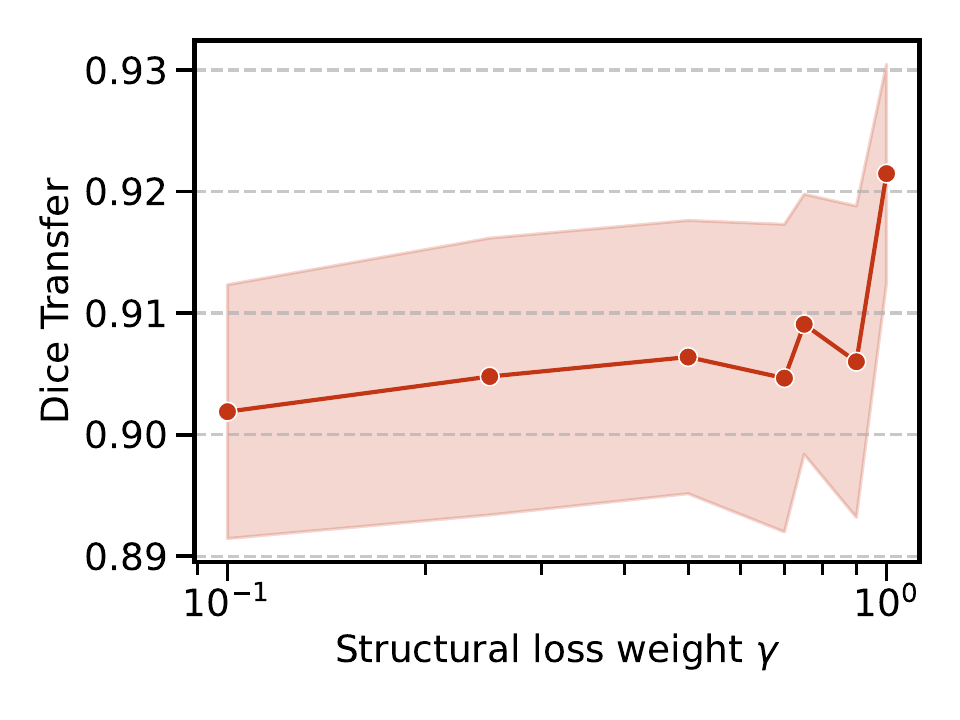} &
    \includegraphics[width=.33\linewidth]{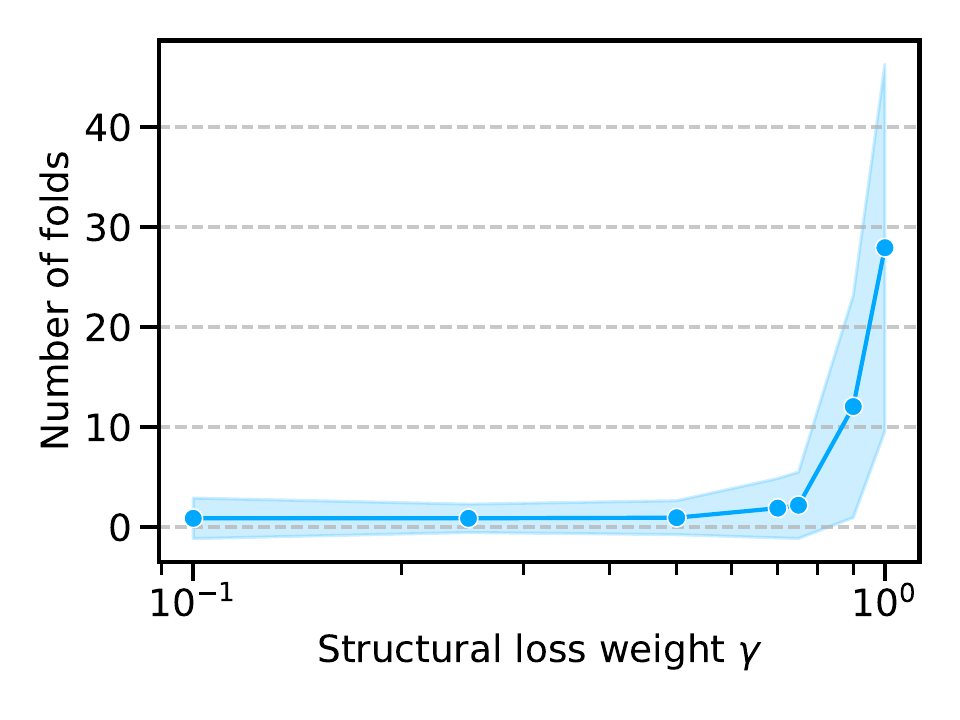} &
    \includegraphics[width=.33\linewidth]{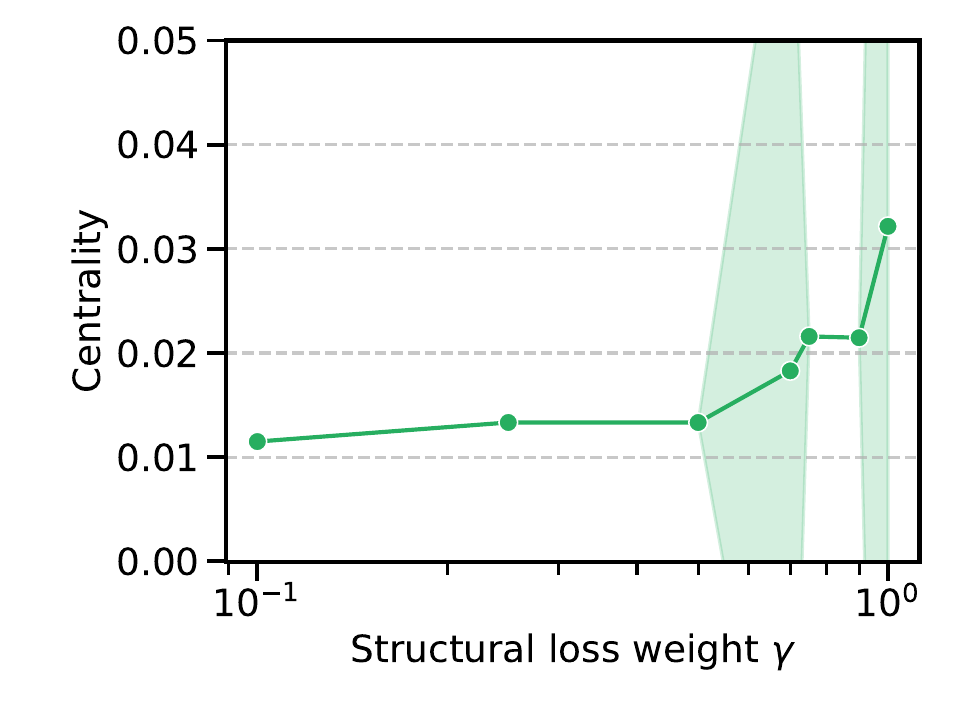}
    \\
    \end{tabular}
    \caption{Hyperparameter sweep over Dice loss weight $\gamma$ with regularization weight $\lambda=1$ on the OASIS-1 validation set. Shaded regions represent one standard deviation from the mean. Plots show the effect on Dice segmentation transfer, number of folded voxels, and Centrality. Our model shows some sensitivity but achieves consistent performance for $\gamma\in[0.1,\ldots,0.7]$. We select $\gamma=0.5$ as it achieves strong segmentation performance while maintaining well-behaved deformation fields.}
    \label{fig:paramsweep-gamma}
\end{figure*}

Figure~\ref{fig:paramsweep-gamma} shows performance while varying $\gamma$ and setting $\lambda=1$. The model shows some sensitivity to the Dice loss weight, though maintains strong performance for $\gamma \in [0.1,\ldots,0.7]$. We select $\gamma=0.5$ and $\lambda=1$ for all experiments in the paper. This set of hyperparameters achieved a reasonable tradeoff between structural matching while maintaining regular and smooth deformation fields.

\section{Additional Qualitative Results}
\begin{figure*}[]
    \centering
    \includegraphics[width=\linewidth]{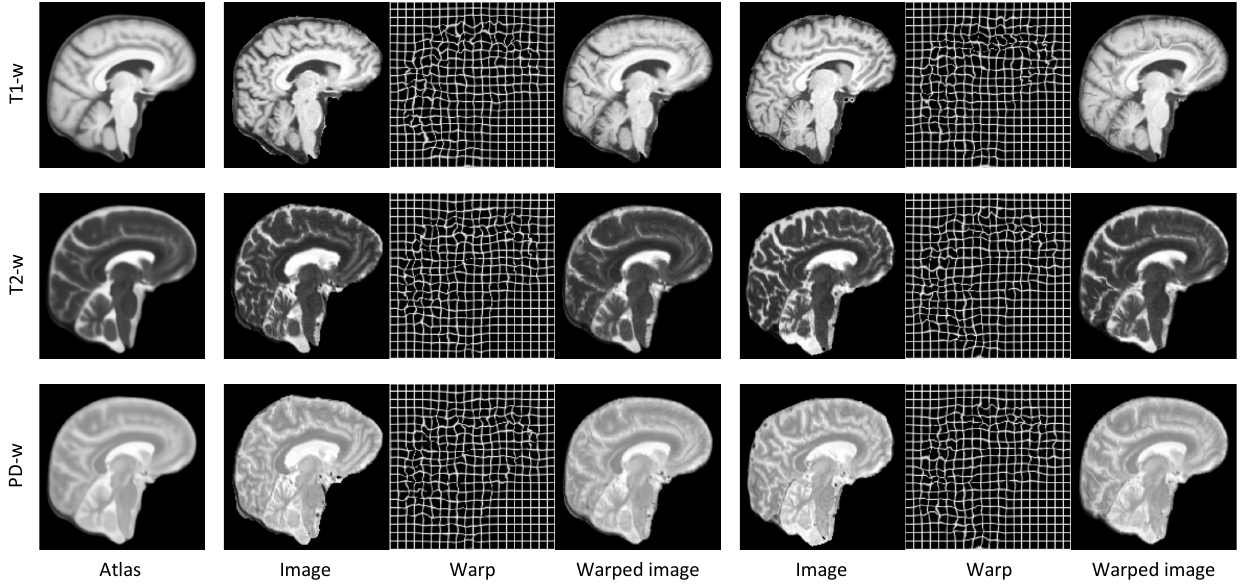}
    \caption{Example images and warps produced by our model on the IXI dataset. }
    \label{fig:images-and-warps}
\end{figure*}

We present additional qualitiative results of our produced atlases. Figure~\ref{fig:images-and-warps} presents example images and warps to the whole-population IXI atlases. Examples are presented for the T1-w, T2-w and PD-w modalities. Despite differences in contrast and image quality, our single model is able to successfully map individual images to the constructed atlases. 

\begin{figure*}
    \centering
    \includegraphics[width=\linewidth]{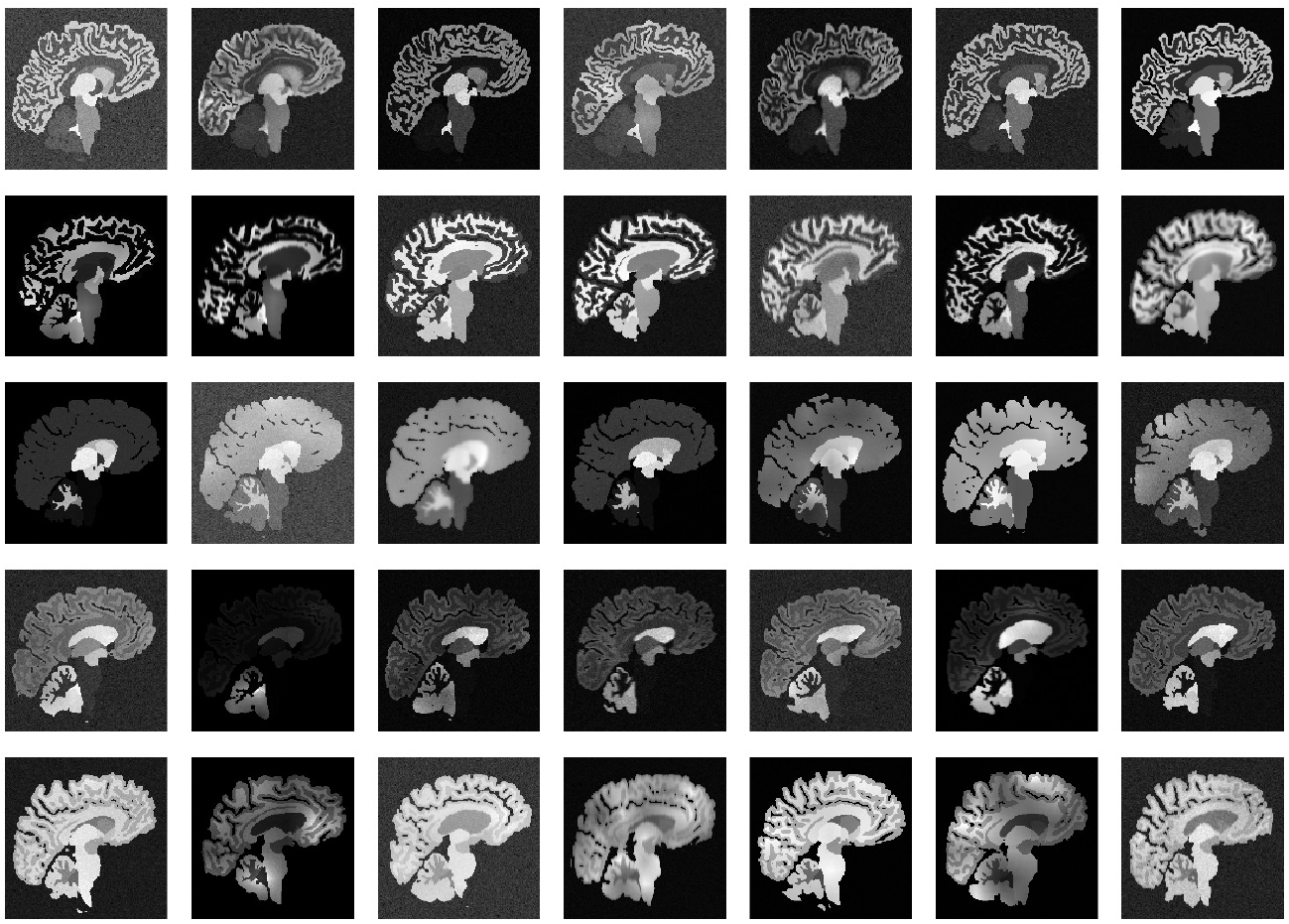}
    \caption{Example synthetic images used in training. Each row represents one group sampled from the same distribution of image contrast, with augmentations performed.}
    \label{fig:example-synth}
\end{figure*}

Figure~\ref{fig:example-synth} presents examples of synthetic images used in training. The variety of imaging contrasts sampled aids our model's ability to generalize to unseen modalities.

\end{document}